\documentclass{article}

\usepackage{array} 
\usepackage{booktabs} 

\usepackage[preprint]{neurips_2026}

\usepackage[utf8]{inputenc} 
\usepackage[T1]{fontenc}    
\usepackage{hyperref}       
\usepackage{url}            
\usepackage{booktabs}       
\usepackage{amsfonts}       
\usepackage{nicefrac}       
\usepackage{microtype}      
\usepackage{xcolor} 
\usepackage{siunitx}
\usepackage{caption}
\usepackage{enumitem}

\usepackage[most]{tcolorbox}
\usepackage{tikz}
\usepackage{xcolor}
\usepackage{tabularx}
\usepackage{subcaption}
\usetikzlibrary{arrows.meta, positioning, shapes.geometric, calc}

\definecolor{gt}{HTML}{D04A1D}
\definecolor{gtbg}{HTML}{FCE9E3}
\definecolor{llm}{HTML}{2E7D32}
\definecolor{llmbg}{HTML}{E6F4EA}
\definecolor{card}{HTML}{F7F7F7}
\definecolor{border}{HTML}{CFCFCF}
\definecolor{muted}{HTML}{777777}

\title{ER-Reason: A Benchmark Dataset for LLM Clinical Reasoning in the Emergency Room}

\author{%
  \textnormal{Nikita Mehandru} \\
  University of California, Berkeley\\
  \texttt{nmehandru@berkeley.edu} \\
  \and
  Niloufar Golchini \\
  University of California, Berkeley\\
  \texttt{ngolchini@berkeley.edu} \\
  \and
  Namrata Garg \\
  University of California, San Francisco\\
  \texttt{namrata.garg@ucsf.edu} \\
  \and
  Kathy T. LeSaint \\
  University of California, San Francisco\\
  \texttt{Kathy.LeSaint@ucsf.edu} \\
  \and
  Christopher J. Nash \\
  Duke University\\
  \texttt{chris.nash@duke.edu} \\
  \and
  Anu Ramachandran \\
  University of California, San Francisco\\
  \texttt{Anu.Ramachandran@ucsf.edu} \\
  \and
  Travis Zack \\
  University of California, San Francisco\\
  \texttt{travis.zack@ucsf.edu} \\
  \and
  Liam G. McCoy \\
  University of Alberta\\
  \texttt{lmccoy@ualberta.ca} \\
  \and
  Adam Rodman \\
  Beth Israel Deaconess Medical Center\\
  \texttt{arodman@bidmc.harvard.edu} \\
  \and
  David Bamman \\
  University of California, Berkeley\\
  \texttt{dbamman@berkeley.edu} \\
  \and
  Melanie Molina* \\
  University of California, San Francisco\\
  \texttt{Melanie.Molina@ucsf.edu} \\
  \and
  Ahmed Alaa* \\
  UC Berkeley and UCSF\\
  \texttt{amalaa@berkeley.edu}
}
\begin{document}

\maketitle

\vspace{-.25in}
\begin{abstract}
\vspace{-.1in}
Existing benchmarks for evaluating the clinical reasoning capabilities of~large~language models (LLMs) often lack a clear definition of ``clinical reasoning''~as~a~construct, fail to capture the full breadth of interdependent tasks within~a~clinical~workflow, and rely on stylized vignettes rather than real-world clinical~documentation. As a result, recent studies have found significant discrepancies~between~LLM~performance on stylized benchmarks derived from medical licensing exams and~their~performance in real-world prospective studies. To address these~limitations,~we~introduce {\small \textsc{ER-Reason}}, a benchmark designed to evaluate LLM reasoning as clinical evidence accumulates across decision-making tasks spanning the full~workflow~of emergency medicine. {\small \textsc{ER-Reason}} comprises 25,174 de-identified clinical notes from 3,437 patients, supporting evaluation across all stages of the emergency department workflow: triage intake, treatment selection, disposition~planning,~and~final diagnosis. Crucially, evaluation in {\small \textsc{ER-Reason}} extends beyond diagnostic accuracy to include stepwise Script Concordance Test (SCT)-style questions grounded in real patient cases, which assess whether LLMs update their diagnostic beliefs in the correct direction and magnitude as clinical evidence accumulates, scored against 2,555 emergency physician annotations. We evaluate reasoning and non-reasoning LLMs on {\small \textsc{ER-Reason}}, and show that our tasks provide a more nuanced view of how LLM reasoning fails on real patient cases than existing benchmarks allow.
\end{abstract}

\vspace{-.1in}

{\bf Code:} \href{https://github.com/AlaaLab/ER-Reason}{https://github.com/AlaaLab/ER-Reason}

{\bf Benchmark dataset:} \href{https://physionet.org/content/er-reason/1.0.0/}{https://physionet.org/content/er-reason/1.0.0/}

\section{Introduction}
\label{Sec1} 
\vspace{-1mm}
While large language models (LLMs) have demonstrated clear value in clinical~documentation~through deployment of tools such as ambient AI scribes \cite{olson2025use, tierney2024ambient, shah2025ambient}, evidence of their utility in supporting clinical decision-making remains mixed. On the one hand, recent work has reported striking~performance~on established reasoning benchmarks; for example, Brodeur et al. \cite{brodeur2026science}, in {\it Science}, showed that OpenAI's o1 model exceeds physician baselines across New England Journal of Medicine (NEJM) clinicopathologic conferences, NEJM Healer cases, and management vignettes. On the other hand, some real-world prospective evaluations paint a more cautious picture. Randomized trials~have~shown~that physicians given access to GPT-4 do not outperform those using conventional resources on diagnostic reasoning \cite{goh2024llm, goh2025gpt4}. Notably, a recent study of OpenAI's consumer-facing ChatGPT Health found that it undertriaged 52\% of gold-standard emergency cases, where it directed patients with~diabetic~ketoacidosis or impending respiratory failure to 24-48 hr evaluation rather than the emergency~department~\cite{ramaswamy2026chatgpt}.

A common criticism of existing studies is the extent to which their experimental design~reflects~the~actual use of LLMs in clinical decision support. For instance, a recent methodological~critique~\cite{navarro2026evaluation}~of~the ChatGPT Health triage study \cite{ramaswamy2026chatgpt} argued that its exam-style format differs fundamentally from how consumers actually interact with medical chatbots, which could undermine the validity of the results. Even Brodeur et al. \cite{brodeur2026science}, who report strong LLM performance on curated patient cases rather than real-world documentation, acknowledge that {\it ``new benchmarks, trials, and technological solutions are required to more faithfully measure clinical encounters.''} This is fundamentally a limitation of \emph{construct validity} \cite{raji2025s, alaa2025medical}; a concern previously raised about early LLM benchmarks that relied on multiple-choice questions from medical licensing exams such as the USMLE \cite{jin2021disease}. 

\begin{figure*}[t]
  \centering
  \includegraphics[width=\textwidth]{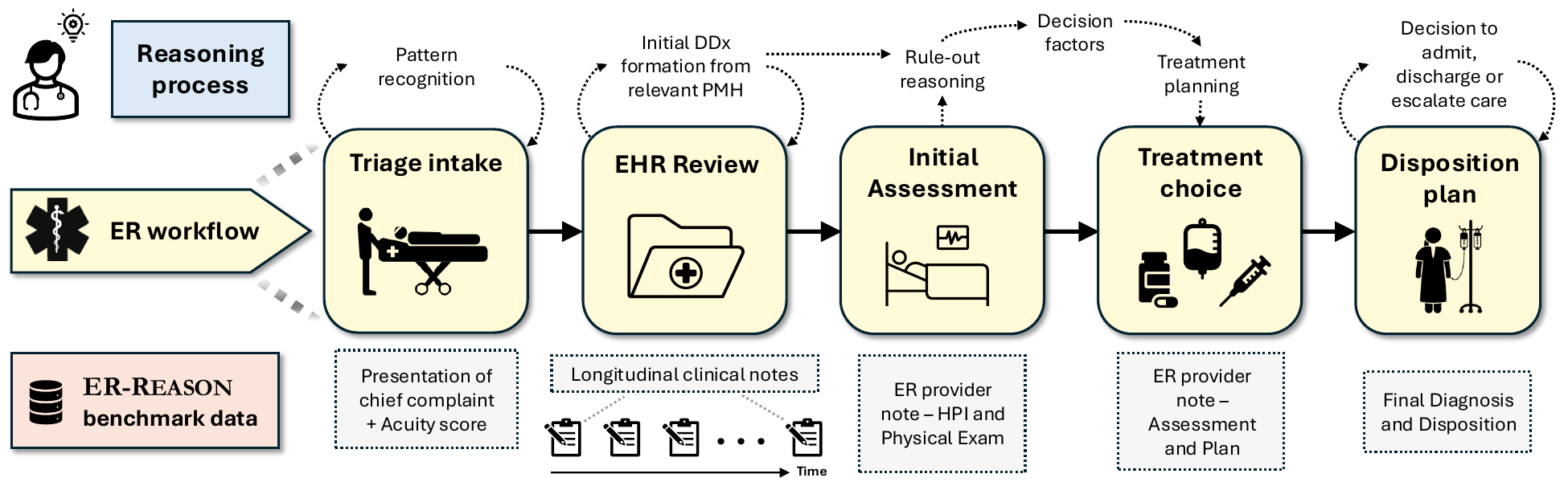}
  \caption{\footnotesize {\bf Overview of the {\small \textsc{ER-Reason}} benchmark.} The benchmark dataset consists of de-identified, longitudinal clinical notes spanning a wide range of document types, including discharge summaries, progress notes, History and Physical exams, consult notes, imaging reports, and ER provider notes. The evaluation tasks cover all key stages of clinical decision-making in the ER workflow, including triage intake, initial assessment, treatment selection, disposition plan, and final diagnosis.}
  \label{fig:1}
  \vspace{2mm}
  \hrule
  \vspace{-3mm}
\end{figure*}

Another limitation in the construct validity of existing benchmarks is that while many claim to evaluate ``clinical reasoning,'' this construct is often conflated with the related capabilities of knowledge recall and diagnostic accuracy. That is, most benchmarks assess a model's final diagnostic decision or its answers to knowledge questions, rather than its ability to revise beliefs under uncertainty as clinical evidence accumulates. Even recent benchmarks that use real-world clinical documentation as input focus on knowledge recall and diagnostic accuracy \cite{wu2025medarena, bedi2026holistic, wornow2023ehrshot}. McCoy et al. \cite{mccoy2025assessment}~show~that the gap between these measurements is empirical and not just conceptual; using Script Concordance Testing, a methodology designed to assess how clinicians revise judgments under uncertainty, they find that LLMs perform considerably worse than on standard multiple-choice benchmarks, which suggests that what licensing exams measure and what clinical reasoning requires are distinct constructs.

{\bf Contribution.} In this paper, we introduce {\small \textsc{ER-Reason}}, an open-access benchmark for~evaluating~the clinical reasoning capabilities of LLMs that prioritizes construct validity. We follow {\it two~design~principles} to address the limitations above. First, we operationalize clinical reasoning as {\it sequential~belief updating}. Following Ledley and Lusted's foundational model of medical decision-making~\cite{ledley1959reasoning},~physicians form an initial differential diagnosis and revise the probability of each hypothesis as new clinical evidence becomes available. Under this view, an LLM demonstrates strong clinical reasoning not only when it reaches the same conclusions as a physician, but when it arrives at them through a similar line of reasoning \cite{gruppen2016clinical}. Second, we design evaluation tasks that mirror realistic decision-making~across~the full workflow of emergency medicine, using de-identified real-world clinical notes as input. 

The emergency room (ER) is a particularly demanding setting for evaluation: physicians must make rapid, high-stakes decisions from fragmented documentation, conditions under which LLMs could support clinicians by synthesizing disparate sources of information. {\small \textsc{ER-Reason}} captures~the~full~journey of patients in the ER of a large academic medical center and comprises the~following~components:

$\bullet$\, \textbf{A large-scale clinical corpus of 25,174 de-identified, longitudinal clinical notes} from 3,984 hospital encounters, spanning discharge summaries, progress notes, history and physical~exams~(H\&Ps), consult notes, imaging and echocardiography reports, and ER provider notes.

$\bullet$\, \textbf{Clinical decision evaluation tasks} across all the key stages of the ER workflow, including triage intake, treatment selection, disposition planning, and final diagnosis (Figure \ref{fig:1}). 

$\bullet$\, \textbf{Script concordance test (SCT)-style reasoning evaluation} on 194 physician-authored~test~questions grounded in real patient cases in {\small \textsc{ER-Reason}} and independently annotated by two ER physicians (2,555 annotations total). We introduce new metrics that operationalize clinical reasoning as sequential belief updating, measuring whether LLMs revise diagnostic beliefs in the correct direction~and~with appropriate magnitude as clinical evidence accumulates, scored against physician consensus.

The clinical corpus of 25K longitudinal notes in {\small \textsc{ER-Reason}} is, to our knowledge, the largest open-access resource of multi-document ER patient trajectories curated for reasoning evaluation, and by spanning the full emergency medicine workflow (from triage to disposition to~diagnosis)~it~captures~interdependencies between clinical decisions that single-stage benchmarks cannot.~The~need~for~such benchmarks has become urgent as consumer-facing health chatbots are now deployed at scale by OpenAI, Anthropic, and others, and prospective human evaluations, though essential, are too slow to keep pace. The recent study in \cite{brodeur2026science}, for instance, evaluated o1-preview, a model~already~superseded~by the time of publication. Automated benchmarks that faithfully measure clinical~reasoning~offer~scalable means of keeping evaluation apace with deployment. The de-identified dataset is available~through~the PhysioNet credentialed-access repository: \href{https://physionet.org/content/er-reason/1.0.0/}{https://physionet.org/content/er-reason/1.0.0/}.

\section{\textsc{ER-Reason}: A Benchmark for LLM Clinical Reasoning in the ER}
\label{Sec3} 

\subsection{Longitudinal Clinical Notes}
\begin{figure}[h]
  \centering
  \begin{minipage}{0.48\textwidth}
    For each ER patient, we include clinical notes from their previous hospital encounter, and thus capture the complexity and temporal progression of patient care across multiple visits. To our knowledge, this is the first publicly released dataset providing longitudinal, multi-encounter clinical notes for ER patients to enable evaluation of LLMs on realistic clinical reasoning tasks. All clinical notes in the \textsc{ER-Reason} dataset were de-identified in accordance with HIPAA Safe Harbor standards to protect patient privacy and were approved by institutional compliance.
  \end{minipage}
  \hfill
  \begin{minipage}{0.5\textwidth}
    \centering
    \includegraphics[width=\textwidth]{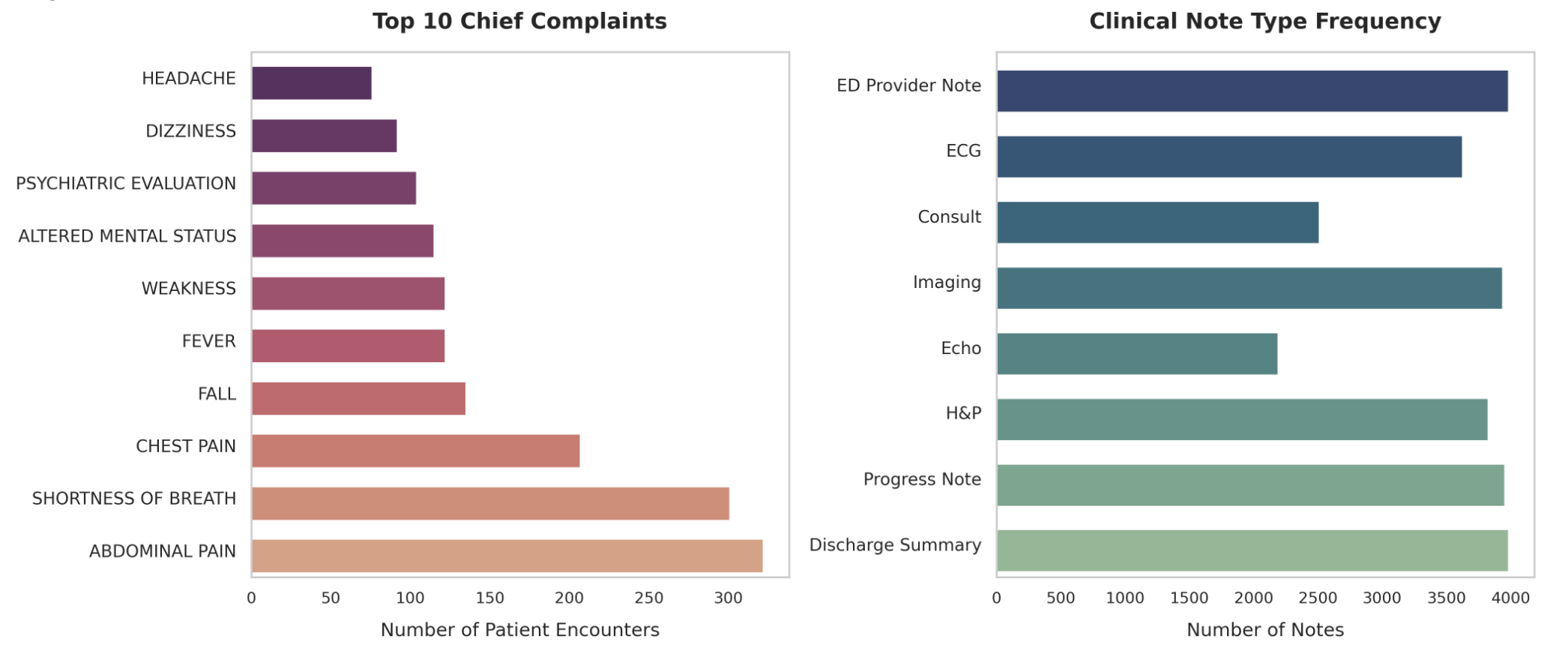}
    \captionof{figure}{{\footnotesize Distribution of the top 10 chief complaints (left) and distribution of note types in~\textsc{ER-Reason}~(right).}}
    \label{fig:cc_notes}
  \end{minipage}
\end{figure}

The \textsc{ER-Reason} dataset consists of longitudinal clinical notes of patients visiting the ER of~a~large academic medical center from March 1, 2022, to March 31, 2024. In total, our~dataset~spans~3,437 patients across 3,984 ER encounters, and encompasses 25,174 de-identified clinical~notes.~Each~patient has multiple unique encounter which contain different note types, such as discharge summaries, progress notes, history and physical examination (H\&P), consults, echocardiogram reports, electrocardiogram (ECG) reports, imaging notes, and ED provider notes. On average, there are 7~notes~per~patient. Unlike standardized note formats such as SOAP (subjective, objective, assessment,~plan)~\cite{wang2024direct}, the diverse and unstructured nature of these notes adds additional complexity for reasoning tasks. The dataset includes 395 unique chief complaints, reflecting the broad range of clinical presentations in the ER; the most frequent complaints were abdominal pain, shortness of breath, and chest~pain~(Figure~\ref{fig:cc_notes}).

\vspace{-0.075in}

\vspace{-.1in}
\subsection{Script Concordance Test (SCT) Physician Annotations}
\vspace{-.05in}
To evaluate clinical reasoning as sequential belief updating, we developed SCT-style questions derived from real patient cases in the ER provider notes in \textsc{ER-Reason}. Each case was authored~by~an~emergency medicine attending physician and presents a clinical scenario alongside three~to~five~differential diagnoses derived from the patient note. New evidence introduced at each step of the process~can~include a new laboratory result, an imaging finding, or an updated physical exam, each of which may increase or decrease the likelihood of one or more the diagnoses on a five-point scale ($-2$ to $+2$). 

In addition to the physician test creators, three emergency medicine physicians reviewed each case: two attending physicians independently annotated each step of the diagnostic~process,~with~the~third~adjudicating disagreements. The resulting consensus scores served as ground truth for all SCT metrics. Across the 194 cases, we collected annotations for 787 SCT items along with~physician-authored~rationales, as well as 194 annotations of the final differential lists, for a total of 2,555 annotations.

{\bf Inter-annotator agreement.} To provide context for our results, we measured agreement between two independent emergency physician annotators; our aim being that model-physician agreement should be at least as strong as physician-physician agreement on the same task. Physicians seldom agreed on the precise ordinal score: physician-physician agreement was 52.6\%, with a Spearman correlation of 0.766 on the final diagnosis. These figures reflect the genuinely challenging nature of sequential diagnostic reasoning and provide essential context for interpreting model performance. Our design parallels that of the landmark study by Brodeur et al., who used two independent physician annotators across their experiments \cite{brodeur2026science}; we extend this with a third physician adjudicator for resolving disagreements. Importantly, our agreement levels are consistent with comparable evaluations~in~this study, which also reported 54\% exact agreement and $\kappa = 0.51$ between physicians scoring differential diagnoses on real-world ER clinical notes. This task is arguably simpler than ours, which requires evaluation across five competing diagnostic hypotheses at multiple decision points. 

\vspace{-.075in}
\begin{figure}[h]
  \centering
  \begin{minipage}[t]{0.55\textwidth}
    \vspace{0pt}
    {\bf Construct validity.} Although our SCT-style evaluation builds on the same psychometric tradition as McCoy et al. \cite{mccoy2025assessment}, it differs in three aspects that are meant to strengthen construct validity. First, our SCT items are sequential within a case rather than independent: each item tests how a model revises its beliefs as new evidence becomes available at the next stage of the ER workflow, mirroring the iterative diagnostic-testing loop introduced by Ledley and Lusted (Figure~\ref{fig:ledley_lusted}) \cite{ledley1959reasoning}. Second, our items are derived from real clinical documentation rather than from existing educational SCT datasets. Third, the evidence accumulated between items corresponds to specific clinical decision points (i.e., triage intake, ER physician evaluation, and admission) rather than to abstract belief-updating. Our goal from these new design choices is to improve the construct validity of our benchmark by measuring clinical reasoning in a workflow-grounded manner that resembles how physicians actually perform the clinical reasoning process.
  \end{minipage}
  \hfill
  \begin{minipage}[t]{0.42\textwidth}
    \vspace{0pt}
    \centering
    \begin{tikzpicture}[
      node distance=7mm,
      evidence/.style={
        draw, rounded corners=2pt, align=center,
        minimum width=38mm, minimum height=10mm,
        font=\scriptsize, inner sep=3pt, fill=blue!8
      },
      revise/.style={
        draw, rounded corners=2pt, align=center,
        minimum width=38mm, minimum height=10mm,
        font=\scriptsize, inner sep=3pt, fill=orange!10
      },
      arrow/.style={-{Stealth[length=2mm]}, thick},
      exit/.style={font=\scriptsize\itshape, align=left}
    ]
      \node[evidence] (test) {Gather new evidence \\ (observe $E_1,\dots,E_n$)};
      \node[revise, below=of test] (diag) {Revise differential \\ (update $r(D_1),\dots,r(D_k)$)};
      \node[evidence, below=of diag] (next) {Determine next tests};

      \draw[arrow] (test) -- (diag);
      \draw[arrow] (diag) -- (next);

      \draw[arrow] (next.west) -- ++(-6mm,0) |- (test.west);

      \draw[arrow] (diag.east) -- ++(8mm,0) node[exit, right] {Final \\ diagnosis};
      \draw[arrow] (next.east) -- ++(8mm,0) node[exit, right] {No further \\ tests};
    \end{tikzpicture}
    \captionof{figure}{{\small Ledley and Lusted's sequential diagnostic testing loop \cite{ledley1959reasoning}: physicians iteratively gather evidence, revise the differential across competing diagnoses, and choose next tests until a final diagnosis is reached. (Figure adapted from Ledley and Lusted's textbook in \cite{ledley1962medical}.)}}
    \label{fig:ledley_lusted}
  \end{minipage}
\end{figure}

\vspace{-.1in}
\subsection{Benchmark Tasks}
\vspace{-.05in}
In this section, we describe the evaluation tasks within \textsc{ER-Reason}.~The~benchmark~consists~of~three tasks, each corresponding to a distinct stage of the ER workflow, as illustrated~in~Figure~\ref{fig:1}.

{\bf $\bullet$\, Clinical decision tasks.} We include {\it three} tasks where clinical reasoning is~evaluated through the accuracy of the model in making correct decisions based on real-world patient documentation,~as~is standard in most existing benchmarks. The first task is {\bf acuity scoring}, which happens at the initial triage intake stage of the ER workflow. This task simulates real-time clinical reasoning that occurs before a full diagnostic workup is available: in triage intake, patient cases are rapidly~evaluated~and prioritized according to their illness severity. We assess the capability of LLMs to determine patient acuity based on clinical presentations, and evaluate the outputted {\it Emergency Severity Index} (ESI) score, which ranges from 1 (critical) to 5 (non-urgent), against the physician-documented ones. 

The second clinical decision task is the {\bf final diagnosis}, which is typically established at the end~of~the ER workflow alongside the disposition plan. Final diagnosis performance is evaluated at two levels:  ICD-10 exact match, which requires the model to predict the precise billing code,~and~Clinical~Classifications Software Refined (CCSR) category accuracy, which maps predictions to one of 530 clinically meaningful categories \cite{malecki2024development}, capturing whether the model reasoned in the right clinical category even when the exact diagnosis code may differ.

Finally, we evaluate {\bf final disposition} in which model decisions~are~compared~against physician-documented ones across ten options: `Discharge', `Admit', `Observation', `Transfer to Another Facility', `AMA', `OR Admit', `LWBS after Triage', `Send~to~L\&D', `Expired', and `Eloped.'

{\bf $\bullet$\, SCT-reasoning tasks.} The middle stages of the ER workflow---EHR review, initial assessment, and treatment choice---are where clinical reasoning is most crucial. Rather than evaluating these stages through free-text generation, \textsc{ER-Reason} operationalizes reasoning at these stages through our SCT-style evaluation. Patient cases were directly derived from the larger dataset, and the task~is~to rule-out reasoning until a final diagnosis is reached. We introduce a new metric, \textit{DxUpdate},~to~measure timestep-level belief update agreement using linearly weighted Cohen's $\kappa$:
\begin{equation}
\kappa_w = 1 - \frac{\sum_{i,j} w_{ij}\, O_{ij}}{\sum_{i,j} w_{ij}\, E_{ij}},
\qquad w_{ij} = \frac{|i - j|}{4},
\end{equation}
where $O_{ij}$ and $E_{ij}$ are the observed and expected counts for physician score $i$ and model score $j$. Weights grow linearly with ordinal distance on the five-point scale ($-2$ to $+2$), so a one-step discrepancy (e.g., $+2$ vs.\ $+1$) incurs proportionally less penalty than a direction-reversing error (e.g., $+2$ vs.\ $-2$). We use linear rather than quadratic weights to preserve proportional penalization across all error magnitudes. DxUpdate is reported as the mean $\kappa_w$ across timesteps.

The DxUpdate metric evaluates agreement between physicians and models on items in the SCT-style test, i.e., how the model processes evidence to update its belief. To evaluate the consequence of those belief updates on the differential diagnosis list for a given patient case we also introduce two metrics, \textit{DxTrajectory} and \textit{FinalDx}, both use Spearman's rank correlation between physician rankings $\mathbf{r}=(r(D_1),\dots,r(D_k))$ and model rankings 
$\hat{\mathbf{r}}=(\hat{r}(D_1),\dots,\hat{r}(D_k))$ over $k$ differentials. DxTrajectory is the mean $\rho$ across all encounter--timestep pairs, and captures whether the model's belief updates produce the correct differential ordering over time. FinalDx is the mean $\rho$ at the final step of each encounter, and captures the accuracy of diagnosis for the patient case given {\it all} evidence.

We evaluate LLMs on each stage individually, but additionally introduce two cross-stage analyses that test whether model outputs form a coherent clinical narrative across the ER workflow. \textbf{Intra-case coherence} measures internal consistency, whether a model's stated belief updates are consistent with its own differential rankings within the SCT evaluation patient set. \textbf{Workflow accuracy} tests whether the model correctly predicts patient acuity, final diagnosis, and final disposition across the full patient cohort. These analyses operationalize a core assumption of ER-Reason: evaluating tasks in isolation is insufficient in detecting clinically dangerous failures in model reasoning.

\section{Related Work}
\label{Sec2}

{\bf Medical Benchmarks for LLMs.} Early efforts to evaluate LLMs in medicine relied on multiple choice question answering (QA) from medical licensing exams, including MedQA \cite{jin2021disease} and PubMedQA \cite{jin2019pubmedqa}, which assess factual recall and basic clinical knowledge.~With~growing~realization~that these benchmarks do not capture true clinical utility \cite{alaa2025medical}, the community moved toward real-world clinical data: EHRSHOT \cite{wornow2023ehrshot} evaluates prediction tasks from structured EHR~data;~MedHelm~\cite{bedi2026holistic} assesses nine frontier models across 37 medical use cases spanning clinical decision support, documentation, and administration; MedArena \cite{wu2025medarena} enables multi-turn dialogue evaluations and head-to-head model comparisons. Despite these advancements, most existing benchmarks measure the ``what'' of clinical decisions—what tests were ordered, what diagnoses were given, what treatments were prescribed, while largely neglecting the ``why'':  the reasoning process underlying~those~decisions. 

{\bf Benchmarks for LLM-based Clinical Reasoning.} A small number of recent benchmarks have started to focus explicitly on evaluating clinical reasoning in LLMs. DiReCT \cite{wang2024direct} evaluates diagnostic reasoning from clinical notes, but covers a limited disease scope and does not capture the multi-hypothesis, sequential structure of real-time environments like the~ER.~Similarly,~Dr. Bench \cite{gao2023dr} covers six tasks built from ten public datasets, but these are centered around reasoning in the~context~of clinical text understanding and diagnosis, and are not grounded in a real-world clinical workflow. MedCaseReasoning \cite{wu2025medcasereasoning} extracts reasoning chains from published case reports and asks whether models mention the right reasons, but not whether models update their beliefs like a physician would as evidence accumulates. While recent work has begun to operationalize sequential diagnosis as an iterative process~\cite{nori2025sequential}, most existing methods still only evaluate final diagnostic accuracy, rather than assessing whether intermediate belief updates are calibrated to physician judgments.

{\bf Script Concordance Tests.} The Script Concordance Test (SCT) was developed in medical education to assess clinical reasoning under uncertainty \cite{charlin2000script, lubarsky2013script}: given a clinical scenario, a hypothesis, and a new piece of evidence, it asks how should a clinician's belief in that hypothesis change? Examinees respond on a five-point ordinal scale from much less likely ($-2$) to much more likely ($+2$), scored against a panel of expert physicians. The SCT captures something multiple-choice tests cannot: the ability to update beliefs appropriately in response to ambiguous or unexpected information. McCoy et al. \cite{mccoy2025language} recently adapted SCTs as a computational benchmark for LLMs, evaluating four frontier models across 750 questions from ten medical datasets, and found models consistently underperform expert clinicians. However, their evaluation assesses one hypothesis at a time. \textsc{ER-Reason} extends this framework by requiring simultaneous belief updating across a full differential diagnosis list of five competing hypotheses, grounded in real ER patient data with prospective annotations from three ER physicians. Importantly, while traditional SCT scoring requires a large panel of physicians, we use three emergency physicians due to the specialized expertise and annotation burden required.

\section{Experiments}
\label{Sec4}

\subsection{Setup}
We evaluate eight LLMs spanning non-reasoning and reasoning  models with both open and closed weights. Our baselines include: DeepSeek-R1, o4-mini, Gemini 2.5 Flash Thinking, GPT-5.2 Thinking, Claude 4.5 Thinking, Phi-4, GPT-5.2, and Claude Sonnet 4.5. For {\bf clinical decision tasks}, we evaluate models under a zero-shot (ZS) setting. We also include a step-back prompting~of~all~baselines \cite{zheng2023take}, which instructs the model to first abstract the patient presentation to higher-level~clinical~principles before committing to a decision, and include these results in Appendix~\ref{app:stepback}.

\vspace{-.15in}
\begin{table}[htbp]
\centering
\small
\caption{{\small Zero-shot performance of models on \textsc{ER-Reason} clinical decision tasks. All metrics are accuracy. 95\% Wilson confidence intervals reported.}}
\vspace{.05in}
\label{tab:main_results_zeroshot}
\resizebox{\textwidth}{!}{%
\begin{tabular}{p{41mm}lcccc}
\toprule
\textbf{Model} & \textbf{Acuity Scoring} & \textbf{Final Diag.} & \textbf{Final Diag. (CCSR)} & \textbf{Final Disp.} \\
\midrule
\multicolumn{5}{l}{\itshape Reasoning Models} \\
\midrule
DeepSeek-R1                & 61.36 {\tiny (59.6--62.6)} & 29.59 {\tiny (28.2--31.0)} & 31.72 {\tiny (30.1--33.0)} & 68.66 {\tiny (67.1--70.0)} \\
o4-mini                    & 61.04 {\tiny (59.5--62.5)} & 35.44 {\tiny (34.0--36.9)} & 36.74 {\tiny (35.1--38.1)} & 69.63 {\tiny (68.4--71.3)} \\
Gemini 2.5 Flash           & 60.27 {\tiny (58.7--61.8)} & 29.99 {\tiny (28.6--31.4)} & 34.49 {\tiny (32.9--35.8)} & 66.62 {\tiny (65.1--68.1)} \\
GPT-5.2 Thinking  & 62.50 {\tiny (61.0--64.0)} & 39.91{\tiny (38.4--41.4)} & 45.09 {\tiny (43.3--46.4)} & 71.49 {\tiny (70.1--72.9)} \\
Claude Sonnet 4.5 Thinking & 59.70 {\tiny (58.2--61.2)} & 35.69 {\tiny (34.2--37.2)} & 41.66 {\tiny (39.9--43.0)} & 67.17 {\tiny (65.7--68.6)} \\
Phi-4                      & 43.12 {\tiny (41.6--44.7)} & 19.28 {\tiny (18.1--20.5)} & 21.55 {\tiny (20.2--22.7)} & 57.52 {\tiny (56.0--59.0)} \\
\midrule
\multicolumn{5}{l}{\itshape Non-Reasoning Models} \\
\midrule
GPT-5.2 &63.15 {\tiny (61.6--64.6)} & 37.73 {\tiny (36.2--39.2)} & 43.25 {\tiny (41.5--44.6)} & 64.76 {\tiny (63.3--66.2)} \\
Claude Sonnet 4.5 & 62.08 {\tiny (60.4--63.4)} & 34.76 {\tiny (33.3--36.3)} & 40.65 {\tiny (38.9--42.0)} &68.27 {\tiny (66.8--69.7)} \\
\bottomrule
\end{tabular}%
}
\end{table}

For the {\bf SCT reasoning tasks}, we conduct four experiments as described below. 
\vspace{-.1in}
\begin{itemize}[leftmargin=1em]
\item First, to disentangle medical knowledge from the clinical reasoning cability, we establish a \textit{clinical knowledge (CK)} baseline. To this end, we measure the model's ability to identify~the~correct~diagnosis when given the one-line patient summary, chief complaint, evidence across all~timesteps,~and~a list of five possible diagnoses. This structure is similar to standard multiple choice~medical~benchmarks \cite{jin2019pubmedqa}, but importantly, constructed on noisy real-world clinical data. 
\end{itemize}

\vspace{-.1in}
We then evaluate three prompting conditions. 
\vspace{-.1in}
\begin{itemize}[leftmargin=1em]
\item In zero-shot, the model receives the patient case, the set of potential differential diagnoses, and a new piece of evidence (e.g., imaging, new lab result) at each timestep, and is asked to assign an ordinal score of how the new evidence affects the likelihood of that differential diagnosis as well as produce a ranked list relative to the other potential differentials. 
\end{itemize}

\vspace{-.1in}
We also add two oracle conditions using our collected physician annotations to test how model performance benefits from human guidance. These rationales do not introduce new clinical information, but instead explain \textit{why} specific evidence should or should not alter the differential. 
\vspace{-.1in}
\begin{itemize}[leftmargin=1em]
\item In \textbf{single oracle (SO)}, the model additionally receives the physician's rationale for that timestep. For example, the model would receive \textit{``tachycardia is a nonspecific sign that can also reflect fever or dehydration’’} or \textit{``pulmonary embolism is not classically associated with abdominal tenderness, making it the least likely diagnosis,’’}, supplementing the evidence with the physician's reasoning. 
\item In \textbf{full oracle (FO)}, the model receives all physician rationales for all prior timesteps cumulatively. At timestep $t$, the model would have access to rationales from timesteps $1$ through $t$. 
\end{itemize}

\vspace{-.05in}
Full details of model inputs, targets, and prompt templates for each task are provided in Appendix~\ref{app:tasks}. 

\subsection{Clinical Decision Tasks}
Table~\ref{tab:main_results_zeroshot} reports performance on acuity assessment, final diagnosis, and disposition planning. We observe that final diagnosis is the hardest task across all models due to its open-ended format: the best zero-shot ICD-10 exact match is 39.91\% (GPT-5.2 Thinking), meaning all models predict the wrong diagnosis more than half the time under realistic clinical conditions. CCSR category accuracy is substantially higher, with GPT-5.2 Thinking reaching 45.09\% accuracy, indicating that larger models reason in the right clinical territory more often than exact-match accuracy suggests, but still fall short of reliable diagnostic performance. This lower absolute accuracy on open-ended clinical tasks is consistent with broader findings. For example, frontier models achieved only 32\% accuracy on HealthBench Hard \cite{arora2025healthbench}, a curated subset of 1,000 realistic health conversations.

\begin{table}[t]
\centering
\small
\caption{{\small Performance in SCT reasoning tasks. (95\% confidence intervals reported from 1,000 bootstraps.)}}
\vspace{.05in}
\label{tab:sct_timestep}
\begin{tabular}{>{\raggedright\arraybackslash}p{38mm} c ccc ccc}
\toprule
 & \multicolumn{3}{c}{\textbf{DxUpdate ($\kappa$)}}
& \multicolumn{3}{c}{\textbf{DxTrajectory ($\rho$)}} \\
\cmidrule(lr){2-4} \cmidrule(lr){5-7}
\textbf{Model}  & \textbf{ZS} & \textbf{SO} & \textbf{FO}
              & \textbf{ZS} & \textbf{SO} & \textbf{FO} \\
\midrule
DeepSeek-R1   & 0.672 {\tiny (0.648--0.694)} & 0.795  & 0.783 & 0.613 {\tiny (0.581--0.638)} & 0.770  & 0.801 \\
o4-mini & 0.691 {\tiny (0.664--0.713)} & 0.809 & 0.807 & 0.609 {\tiny (0.579--0.641)} & 0.786 & 0.793 \\
Gemini 2.5 Flash    & 0.657 {\tiny (0.619--0.676)} & 0.762 & 0.747 & 0.613 {\tiny (0.603--0.663)} & 0.738 & 0.782 \\
GPT-5.2 Thinking    & 0.676 {\tiny (0.647--0.702)} & 0.796 & 0.785 & 0.638 {\tiny (0.606--0.669)} & 0.769 & 0.801 \\
Claude 4.5 Thinking & 0.652 {\tiny (0.624--0.676)} & 0.756 & 0.742 & 0.591 {\tiny (0.560--0.625)} & 0.719 & 0.780 \\
Phi-4  & 0.538 {\tiny (0.501--0.570)} & 0.719 & 0.691 & 0.525 {\tiny (0.491--0.559)} & 0.703 & 0.741 \\
\midrule
GPT-5.2  & 0.677 {\tiny (0.646--0.701)} & 0.795 & 0.789 & 0.646 {\tiny (0.618--0.680)} & 0.780 & 0.813 \\
Claude Sonnet 4.5   & 0.662 {\tiny (0.632--0.682)} & 0.771 & 0.758 & 0.636 {\tiny (0.605--0.669)} & 0.743 & 0.792 \\
\bottomrule
\end{tabular}
\end{table}

\vspace{-.1in}
\begin{table}[htbp]
\centering
\small
\caption{{\small Final diagnosis performance on the \textsc{ER-Reason} SCT evaluation. CK = clinical knowledge baseline (all evidence provided simultaneously); Acc.\ = top-1 accuracy (\%); $\rho$ = Spearman rank correlation against physician consensus. ZS = zero-shot, SO = single oracle, FO = full oracle.}}
\vspace{.05in}
\label{tab:sct_finaldx}
\begin{tabular}{>{\raggedright\arraybackslash}p{38mm} c cc cc cc}
\toprule
\textbf{Model} & \textbf{CK} & \multicolumn{2}{c}{\textbf{ZS}} & \multicolumn{2}{c}{\textbf{SO}} & \multicolumn{2}{c}{\textbf{FO}} \\
\cmidrule(lr){2-2} \cmidrule(lr){3-4} \cmidrule(lr){5-6} \cmidrule(lr){7-8}
 & \textbf{Acc.} & \textbf{Acc.} & \textbf{$\rho$} & \textbf{Acc.} & \textbf{$\rho$} & \textbf{Acc.} & \textbf{$\rho$} \\
\midrule
DeepSeek-R1  &56.2& 62.4 {\tiny(56.4–69.9)} & 0.562 & 75.8 & 0.738 & 81.4& 0.814\\
o4-mini    &58.8& 65.5 {\tiny(58.9–72.1)}& 0.610 & 80.4 & 0.775 & 77.3 & 0.789 \\
Gemini 2.5 Flash &53.6& 68.0 {\tiny(61.2–74.2)} & 0.601 & 76.3 & 0.703 & 79.9 & 0.796 \\
GPT-5.2 Thinking    &61.9& 70.1 {\tiny(63.3–76.1)} & 0.627 & 85.6 & 0.791 & 88.1 & 0.841 \\
Claude 4.5 Thinking   &62.4& 64.2 {\tiny(57.3–70.7)} & 0.573& 78.9& 0.705& 83.5& 0.795\\
Phi-4  &53.1& 49.5 {\tiny(43.0–57.0)} & 0.443 & 70.6 & 0.677 & 77.3 & 0.742 \\
\midrule
GPT-5.2    &61.3& 69.6 {\tiny(62.8–75.6)} & 0.634 & 82.5 & 0.781 & 85.1 & 0.842 \\
Claude Sonnet 4.5   &61.3& 63.4 {\tiny(56.2–69.7)} & 0.602 & 79.9 & 0.720 & 85.1 & 0.803 \\

\bottomrule
\end{tabular}
\end{table}

\subsection{SCT Reasoning Tasks}
{\bf Limitations of LLMs in SCT sequential reasoning.} We observe significant improvement~in~DxUpdate and DxTrajectory from zero-shot to single and full oracle, with DxTrajectory showing~larger~and more consistent gains than DxUpdate, indicating that ranking the full differential list benefits more from reasoning scaffolding than assigning ordinal update magnitudes at individual timesteps (Table~\ref{tab:sct_timestep}). The performance gap between zero-shot and oracle conditions may suggest limitations in how models structure and update their clinical beliefs rather than their access to medical knowledge alone. 

{\bf Reasoning models do not excel in sequential SCT-style tasks.} Notably, reasoning models do not outperform non-reasoning ones. We did not find inference-time chain-of-thought to provide benefit on sequential belief updating: GPT-5.2 Thinking shows no significant difference from GPT-5.2 on either metric (DxUpdate: $p = 0.455$; DxTrajectory: $p = 0.192$), and Claude 4.5 Thinking significantly underperforms its base model on DxTrajectory ($\Delta\rho = 0.044$, 95\% CI: $0.022$--$0.064$, $p < 0.001$). This contrasts with recent work evaluating LLMs on sequential clinical vignettes with fixed answer keys, where reasoning models show significant advantages \cite{rao2026large}. The discrepancy raises questions about whether reasoning models' apparent strengths hold up outside structured textbook problems, particularly in the kind of sequential, evidence-based reasoning that real clinical judgment demands.  

{\bf Compounding diagnostic errors due to errors in sequential reasoning.} Figure~\ref{fig:figure3} shows top-1 diagnostic accuracy across timesteps (i.e., sequential SCT items within a patient case),~stratified~by the total number of SCT items. We observe that accuracy degrades as evidence accumulates across all models, likely due to error propagation from miscalibrated differential rankings at earlier timesteps. Table~\ref{tab:sct_finaldx} shows that both top-1 accuracy and Spearman $\rho$ improve consistently with the addition of physician reasoning from ZS to SO to FO across all models. The consistent improvement from SO to FO demonstrates that physician rationales provide compounding benefit as evidence accumulates, and that zero-shot under-performance may reflect the model's struggle to reason over sequential evidence. We note that oracle rationales may improve performance partly by signaling salience, specifically directing the model's attention to the most relevant differential, rather than purely improving reasoning structure. This effect may be most pronounced in patient cases with fewer competing hypotheses.

\begin{figure}[t]
    \centering
    \includegraphics[width=1\linewidth]{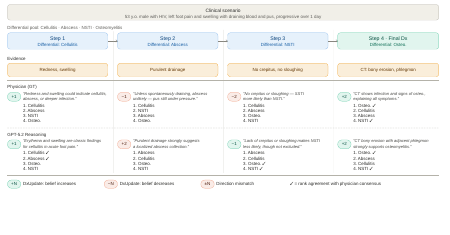}
    \vspace{-.65in}
    \caption{{\footnotesize Sample SCT case from {\small \textsc{ER-Reason}} illustrating sequential belief updates by a physician~(GT)~and GPT-5.2 Thinking across four evidence timesteps. GPT-5.2 reaches the correct final diagnosis,~but~exhibits~incorrect~intermediate~reasoning. Each step shows the differential hypothesis under consideration, new clinical evidence introduced at that timestep, and physician and model rationales. DxUpdate score (teal = belief increase, coral = belief decrease, red = direction mismatch with physician), and DxTrajectory ranking. \checkmark denotes rank agreement with physician consensus.}}
    \label{fig:placeholder}
\end{figure}

\begin{figure}[t]
    \centering
    \vspace{-.15in}
    \includegraphics[width=1\linewidth]{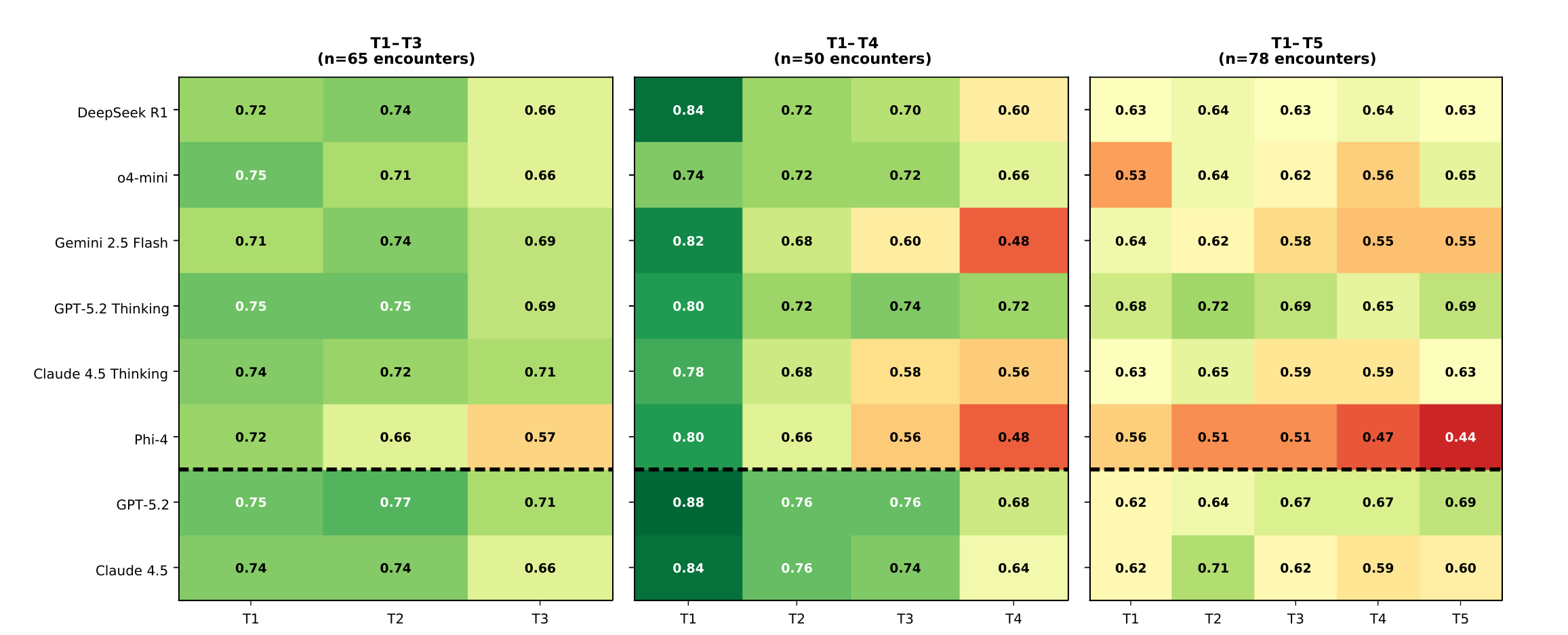}
    \caption{{\small Top-1 diagnostic accuracy across accumulating SCT items in our SCT-style reasoning evaluation.}}
    \label{fig:figure3}
    \vspace{2mm}
    \hrule
    \vspace{-5mm}
\end{figure}

{\bf Consistency of reasoning and diagnosis.} Table~\ref{tab:intra_case} shows that most models maintain high intra-case coherence: when a model assigns a non-zero belief update, it almost always ranks that diagnosis in the corresponding direction. GPT-5.2 achieves perfect coherence (100\%), followed by Gemini 2.5 Flash (99.7\%) and o4-mini (99.5\%). Claude 4.5 Thinking (N=737) and Phi-4 (N=672) assign non-zero updates far more often than conservative models such as Gemini 2.5 Flash (N=318) and GPT-5.2 Thinking (N=323), yet are the least internally consistent at 87.4\% and 91.8\% respectively. Notably, Claude 4.5 Thinking assigns nearly twice as many non-zero updates as its base model Claude Sonnet 4.5 (N=418) while achieving lower coherence ($87.4\%$), suggesting that extended chain-of-thought reasoning may increase directional commitment without a corresponding gain in consistency. More broadly, models making the most directional claims are the least internally consistent. 

\begin{table}[t]
\centering
\small
\caption{{\footnotesize Internal reasoning consistency on 194 SCT cases (n=787). We measure the proportion of non-zero diagnosis timestep pairs where the sign of the model's belief update matches the direction of rank change. N reports the number of non-zero update pairs per model; low N indicates a tendency to abstain from directional claims.}}
\label{tab:intra_case}
\vspace{.05in}
\begin{tabular}{lcc}
\toprule
\textbf{Model} & \textbf{Coherence Rate (CI)}& \textbf{N (non-zero pairs)} \\
\midrule
DeepSeek-R1          & 98.4 {\tiny (96.5--99.3)} & 374 \\
o4-mini              & 99.5 {\tiny (98.3--99.9)} & 429 \\
Gemini 2.5 Flash     & 99.7 {\tiny (98.2--99.9)} & 318 \\
GPT-5.2 Thinking     & 98.5 {\tiny (96.4--99.3)} & 323 \\
Claude 4.5 Thinking  & 87.4 {\tiny (84.8--89.6)} & 737 \\
Phi-4                & 91.8 {\tiny (89.5--93.7)} & 672 \\
\midrule
GPT-5.2              & 100.0 {\tiny (98.9--100.0)} & 354 \\
Claude Sonnet 4.5    & 97.1 {\tiny (95.0--98.4)} & 418 \\
\bottomrule
\end{tabular}
\vspace{-.2in}
\end{table}

\begin{table}[t]
\centering
\footnotesize
\caption{{\footnotesize Cross-stage workflow accuracy ($n=3{,}285$; 2,258 high-acuity, 1,027 low-acuity). Accuracy (\%) reports the proportion of patients where predicted disposition and diagnosis are jointly correct, stratified by acuity bucket. High-acuity includes ESI 1--2 and ESI 3 patients who were admitted; low-acuity includes ESI 4--5 and ESI 3 patients who were discharged. Discordant acuity--disposition pairs (e.g., high-acuity patients discharged), non-admit/discharge dispositions (transfer, AMA, eloped), and unspecified acuity are excluded ($n=699$). 95\% Wilson confidence intervals reported.}}
\label{tab:cross_stage}
\vspace{.05in}
\begin{tabular}{l c c}
\toprule
\textbf{Model}
& \textbf{ESI 1--2 + ESI 3 (Admitted)} & \textbf{ESI 4--5 + ESI 3 (Discharged)}\\
& \textbf{Admit $\cap$ CCSR Inpatient}
& \textbf{Discharge $\cap$ CCSR Outpatient} \\
\midrule
DeepSeek-R1          & 73.03 {\tiny(71.16--74.82)} & 33.98 {\tiny(31.15--36.93)} \\
o4-mini              & 74.76 {\tiny(72.92--76.51)} & 37.59 {\tiny(34.67--40.59)} \\
Gemini 2.5 Flash     & 85.87 {\tiny(84.37--87.25)} & 27.85 {\tiny(25.19--30.67)} \\
GPT-5.2 Thinking    & 80.29 {\tiny(78.60--81.88)} & 51.70 {\tiny(48.65--54.75)} \\
Claude 4.5 Thinking  & 86.09 {\tiny(84.61--87.46)} & 33.11 {\tiny(30.30--36.04)} \\
Phi-4                & 71.97 {\tiny(70.08--73.78)} & 8.86 {\tiny(7.27--10.76)} \\
\midrule
GPT-5.2              & 87.42 {\tiny(85.99--88.73)} & 37.00 {\tiny(34.10--40.00)} \\
Claude Sonnet 4.5    & 85.43 {\tiny(83.91--86.82)} & 37.29 {\tiny(34.39--40.29)} \\
\bottomrule
\end{tabular}%
\vspace{-.2in}
\end{table}

\vspace{-.05in}
\subsection{Evaluation Across the ER Workflow}
Finally, we evaluate the model performance across the entire ER workflow by accounting for the interconnected nature of clinical decisions across all stages. Table~\ref{tab:cross_stage} shows the rates by which a model assigns a patient the correct acuity score and then makes the right disposition plan for triaged patients. We found this cross-stage workflow accuracy to be substantially higher for high-acuity patients (ESI 1--2 + admitted ESI 3) than for low-acuity ones (ESI 4--5 + discharged ESI 3). Base rates and exclusion criteria for this analysis are reported in Appendix Table~\ref{tab:base_rates}. GPT-5.2 leads on the high-acuity cohort (87.42\%), while GPT-5.2 Thinking leads on the low acuity cohort (50.14\%), the only model to exceed 50\% on discharged patients. This systematic overtriage pattern may indicate that models are better calibrated for the ``worst-first'' reasoning pattern. Notably, Gemini 2.5 Flash achieves the second-highest high-acuity accuracy (85.87\%), yet one of the lowest low-acuity scores (27.85\%), illustrating how strong per-task performance can mask failures in cross-stage performance.

\section{Conclusion}
Existing benchmarks fail to capture clinical reasoning, because they neither define it as a construct nor evaluate it on real-world data. {\small \textsc{ER-Reason}} addresses both by grounding evaluation in Ledley and Lusted's sequential diagnostic-testing framework to define what clinical reasoning is,~and~instantiating it on real-world emergency medicine documentation rather than curated vignettes. We hope {\small \textsc{ER-Reason}} provides a methodological template for future evaluation of clinical AI in realistic settings. 

{\bf Limitations.} We did not control or vary the inference-time thinking budget for extended chain-of-thought models, so future work could examine whether increasing the token budget allocated to reasoning models recovers performance. While the oracle experiments establish an upper bound on what physician-style reasoning scaffolding can recover, they do not indicate how to close this gap, whether through fine-tuning on clinical reasoning traces, reinforcement learning from physician feedback, or improved prompting strategies. Finally, our benchmark does not yet account for hospital-level contextual factors, such as bed availability, staffing constraints, and institutional protocols, which can significantly influence decisions such as triage prioritization and patient disposition.

\bibliographystyle{unsrt}
\bibliography{ref}

\appendix
\newpage
\section{Technical appendices and supplementary material}
\subsection{Benchmark Comparisons}
\label{app:comparison}
\begin{table}[htbp]
\centering
\caption{Comparison between benchmark datasets for clinical LLMs. }
\label{tab:benchmark_comparison}
\renewcommand{\arraystretch}{1.2} 
\resizebox{\textwidth}{!}{%
\begin{tabular}{llllll}
\toprule
\textbf{Benchmark} &  \textbf{\# of Patients} & \textbf{Note Types} & \textbf{\# of Notes} & \textbf{Clinical Setting} & \textbf{Reasoning} \\
\midrule
\textbf{\textsc{ER-Reason}} &  3,437 & \begin{tabular}[c]{@{}l@{}}Discharge Summary, Progress, Consult,\\ H\&P, Imaging, Echo, ECG, ED Provider\end{tabular} & 25,174 & ER & Yes \\
&  &  &  &  &    \\[-2ex]
\hline  
&  &  &  &  &    \\[-2ex]
DiRECT \cite{wang2024direct} & -- & Free-text, SOAP-style notes & 511 & Hospital \& ICU & Yes \\
Xie et al. \cite{xie2022benchmarking} & 217k & Free-text notes, vital signs & -- & ER & No \\
MedCaseReasoning \cite{wu2025medcasereasoning} & -- & Case reports (PMC) & 14,489 QA & Outpatient/Case Reports & Yes \\
DR. BENCH \cite{gao2023dr} & -- & Progress Notes & 603 & ICU &  Yes \\
HiRID-ICU \cite{hugo2021hiridicu} & 33,905 & EHR codes, vital signs & -- &  ICU & No \\
PROBSUM \cite{gao2023overview} & -- & Progress notes & 1k & Hospital & No \\
EHRSHOT \cite{wornow2023ehrshot} & 6,739 & \begin{tabular}[c]{@{}l@{}} Tabular EHR data\end{tabular} & None \begin{tabular}[c]{@{}l@{}}--\end{tabular} & \begin{tabular}[c]{@{}l@{}}Hospital\end{tabular} & No \\
MC-BEC \cite{chen2023multimodal} & 63k & Radiology reports & -- & ER & No \\
MEDALIGN \cite{fleming2024medalign} & 276 & \begin{tabular}[c]{@{}l@{}} Structured EHR data \end{tabular} & \begin{tabular}[c]{@{}l@{}}983 instructions\\(303 with responses)\end{tabular} & \begin{tabular}[c]{@{}l@{}}Mixed inpatient\\and ambulatory\end{tabular} & No \\
\bottomrule
\end{tabular}%
}
\end{table}

\subsection{Step-Back Prompting Results}
\label{app:stepback}

\begin{table}[htbp]
\centering
\caption{Step-back performance of models on \textsc{ER-Reason} classification tasks. All metrics are accuracy.}
\label{tab:main_results_stepback}
\resizebox{\textwidth}{!}{%
\begin{tabular}{lcccc}
\toprule
\textbf{Model} & \textbf{Acuity} & \textbf{Final Diag.} & \textbf{Final Diag. (CCSR)} & \textbf{Final Disp.} \\
\midrule
\multicolumn{5}{l}{\itshape Reasoning Models} \\
\midrule
DeepSeek-R1        & 61.85 & 30.37 & 32.48 & 65.23 \\
o4-mini                    & 63.20 & 37.58 & 38.63 & 65.11 \\
Gemini 2.5 Flash           & 60.27 & 32.03 & 37.65 & 62.63 \\
\textbf{GPT-5.2 Thinking}  & \textbf{63.28} & \textbf{44.20} & \textbf{49.18} & \textbf{70.26} \\
Claude Sonnet 4.5 Thinking & 62.47 & 35.14 & 40.65 & 64.88 \\
\midrule
\multicolumn{5}{l}{\itshape Non-Reasoning Models} \\
\midrule
\textbf{GPT-5.2}           & \textbf{62.22} & \textbf{43.65} & \textbf{48.64} & 62.00 \\
\textbf{Claude Sonnet 4.5} & 61.57 & 34.59 & 40.27 & \textbf{66.11} \\
Phi-4                      & 53.94 & 18.28 & 21.47 & 44.14 \\
Qwen3.5-2B                 & 57.10 &  3.71 &  3.52 & 59.54 \\
\bottomrule
\end{tabular}%
}
\end{table}

\subsection{FinalDx Results}
\begin{table}[h]
\centering
\small
\caption{Final diagnosis performance on the \textsc{ER-Reason} SCT evaluation. Acc. = top-1 accuracy (\%); $\rho$ = Spearman rank correlation against physician consensus. ZS = zero-shot, SO = single oracle, FO = full oracle.}
\label{tab:sct_finaldx_appendix}
\begin{tabular}{>{\raggedright\arraybackslash}p{38mm} cc cc cc}
\toprule
& \multicolumn{2}{c}{\textbf{ZS}}
& \multicolumn{2}{c}{\textbf{SO}}
& \multicolumn{2}{c}{\textbf{FO}} \\
\cmidrule(lr){2-3} \cmidrule(lr){4-5} \cmidrule(lr){6-7}
\textbf{Model} & \textbf{Acc.} & \textbf{$\rho$}
              & \textbf{Acc.} & \textbf{$\rho$}
              & \textbf{Acc.} & \textbf{$\rho$} \\
\midrule
DeepSeek-R1           & 62.4 & 0.562 & 75.8 & 0.738 & 81.4& 0.814\\
o4-mini              & 65.5 & 0.610 & 80.4 & 0.775 & 77.3 & 0.789 \\
Gemini 2.5 Flash     & 68.0 & 0.601 & 76.3 & 0.703 & 79.9 & 0.796 \\
GPT-5.2              & 69.6 & 0.634 & 82.5 & 0.781 & 85.1 & 0.842 \\
GPT-5.2 Thinking   & 70.1 & 0.627 & 85.6 & 0.791 & 88.1 & 0.841 \\
Claude Sonnet 4.5    & 63.4 & 0.602 & 79.9 & 0.720 & 85.1 & 0.803 \\
Claude 4.5 Thinking  & 64.2& 0.573& 78.9& 0.705& 83.5& 0.795\\
Phi-4                & 49.5 & 0.443 & 70.6 & 0.677 & 77.3 & 0.742 \\
\bottomrule
\end{tabular}
\end{table}

\subsection {Top-1 Diagnostic Accuracy by Timestep}
\label{app:timestep}
\begin{table}[h!]
\centering
\caption{Top-1 diagnostic accuracy by timestep across all models ($n$ varies by timestep due to differential counts in patient cases).}
\label{tab:top1_by_timestep}
\setlength{\tabcolsep}{6pt}
\renewcommand{\arraystretch}{1.15}
\begin{tabular}{l *{5}{c}}
\toprule
& \multicolumn{5}{c}{\textbf{Top-1 Accuracy by Timestep}} \\
\cmidrule(lr){2-6}
\textbf{Model} & \textbf{T1} & \textbf{T2} & \textbf{T3} & \textbf{T4} & \textbf{T5} \\
& \textit{(n=194)} & \textit{(n=194)} & \textit{(n=193)} & \textit{(n=128)} & \textit{(n=78)} \\
\midrule
Phi-4               & 0.680 & 0.603 & 0.544 & 0.477 & 0.436 \\
Claude Sonnet 4.5   & 0.711 & 0.731 & 0.663 & 0.609 & 0.603 \\
GPT-5.2             & 0.732 & 0.716 & 0.705 & 0.672 & 0.692 \\
\midrule
DeepSeek-R1         & 0.716 & 0.696 & 0.658 & 0.625 & 0.628 \\
Gemini 2.5 Flash    & 0.711 & 0.675 & 0.622 & 0.523 & 0.551 \\
o4-mini             & 0.660 & 0.686 & 0.658 & 0.602 & 0.654 \\
Claude 4.5 Thinking & 0.706 & 0.686 & 0.627 & 0.579 & 0.628 \\
GPT-5.2 Thinking  & 0.737 & 0.732 & 0.705 & 0.680 & 0.692 \\
\bottomrule
\end{tabular}
\end{table}



\section*{Experiment Prompts}
\label{app:tasks}
\addcontentsline{toc}{section}{Experiment Prompts}

The following documents the exact prompts used in each experiment. For all tasks, the \textit{system prompt} is passed as the \texttt{system} field
and the \textit{user prompt} is constructed dynamically per patient via the
\texttt{build\_prompt()} function in the corresponding experiment script. For zero-shot experiments across the three classification tasks (acuity, final diagnosis, and final disposition), the prompts are just the Call 2 without the principles included. 

\subsection*{A.1\quad Acuity Prediction}

\paragraph{System Prompt.}
\begin{quote}
\ttfamily
You are an experienced Emergency Department triage nurse.
\end{quote}

\paragraph{Step-Back Call 1 — Retrieve Principles (user turn).}
\begin{quote}
\ttfamily
What are the key clinical principles and criteria that differentiate each ESI triage level?
Please describe the vital sign thresholds, symptom patterns, chief complaint characteristics,
and risk factors that indicate each of the following acuity levels:
Immediate, Emergent, Urgent, Less Urgent, Non-Urgent.
\end{quote}

\paragraph{Step-Back Call 2 — Per-Patient Prompt (user turn).}
\begin{quote}
\ttfamily
Using the following clinical triage principles:\newline
\{principles\}\newline
\newline
Now assign the acuity level for this patient:\newline
\newline
Age: \{age\}\newline
Sex: \{sex\}\newline
Race: \{firstrace\}\newline
Chief Complaint: \{primarychiefcomplaintname\}\newline
Vital Signs: \{Vital\_Signs\}\newline
\newline
Select the most appropriate acuity level from the following options ONLY:\newline
`Immediate', `Emergent', `Urgent', `Less Urgent', `Non-Urgent'\newline
\newline
Respond with ONLY ONE of these five options. No explanation.
\end{quote}
\subsection*{A.2\quad Disposition Prediction}

\paragraph{System Prompt.}
\begin{quote}
\ttfamily
You are an experienced Emergency Department physician.
\end{quote}

\paragraph{Step-Back Call 1 — Retrieve Principles (user turn).}
\begin{quote}
\ttfamily
What are the key clinical principles and criteria that determine ED disposition?
Please describe what patient factors, diagnoses, vital sign patterns, social circumstances,
and clinical findings typically lead to each of the following outcomes:
Discharge, Admit, Observation, Transfer to Another Facility, AMA, OR Admit,
LWBS after Triage, Send to L\&D, Eloped.
\end{quote}

\paragraph{Step-Back Call 2 — Per-Patient Prompt (user turn).}
\begin{quote}
\ttfamily
Using the following clinical principles:\newline
\{principles\}\newline
\newline
Now predict the most likely ED disposition for this patient:\newline
\newline
Age: \{Age\}\newline
Sex: \{sex\}\newline
Chief Complaint: \{primarychiefcomplaintname\}\newline
Primary ED Diagnosis: \{primaryeddiagnosisname\}\newline
\newline
ED Provider Note:\newline
\{cleaned\_note\}\newline
\newline
Choose ONLY from the following options:\newline
`Discharge', `Admit', `Observation', `Transfer to Another Facility', `AMA',
`OR Admit', `LWBS after Triage', `Send to L\&D', `Eloped'\newline
\newline
Respond with ONLY ONE of these options on the last line. No explanation.
\end{quote}

\noindent\textit{Note:} The ED Provider Note is truncated at the \texttt{"Final
Disposition and ED Course"} section header, and the disposition label is redacted
as \texttt{[REDACTED]} within the note text to prevent label leakage.

\subsection*{A.3\quad Final Diagnosis Prediction}

\paragraph{System Prompt.}
\begin{quote}
\ttfamily
You are an experienced Emergency Department physician.
\end{quote}

\paragraph{Per-Patient Prompt (user turn).}
\begin{quote}
\ttfamily
Predict the most likely primary ED diagnosis for this patient.\newline
\newline
Age: \{Age\}\newline
Sex: \{sex\}\newline
Chief Complaint: \{primarychiefcomplaintname\}\newline
\newline
Past Medical History (Most Recent Discharge Summary):\newline
\{Discharge\_Summary\_Text\}\newline
\newline
ED Provider Note:\newline
\{cleaned\_note\}\newline
\newline
Based on the clinical information above, predict the most likely primary ED diagnosis
for this patient.\newline
Respond with ONLY the ICD-10 code followed by the diagnosis name in this exact format:\newline
{[ICD-10 CODE] [DIAGNOSIS NAME]}\newline
Example: J18.9 Pneumonia, unspecified\newline
No explanation, no alternatives, no additional text.
\end{quote}

\noindent\textit{Note:} The ED Provider Note is truncated at the \texttt{"Final
Disposition and ED Course"} section header. Occurrences of the ground-truth diagnosis
name within the note are replaced with \texttt{[DIAGNOSIS REDACTED]} to prevent
label leakage.

\subsection*{A.4\quad SCT Clinical Reasoning — Baseline}

\paragraph{System Prompt.}
\begin{quote}
\ttfamily
You are an experienced emergency medicine physician reasoning through a clinical case.\newline
You will be given a patient presentation and a fixed list of diagnoses to consider.\newline
At each step, you will receive a new piece of evidence paired with one of those diagnoses.\newline
Your job is to update your belief about that diagnosis based on the evidence, then re-rank
the FULL list.\newline
\newline
You MUST only use the diagnoses from the fixed list provided. Do not add, rename, or invent
any diagnoses.\newline
\newline
Respond ONLY in the following JSON format. Do not include any explanation outside the JSON.\newline
\{\newline
\quad"dx\_updates": [\newline
\quad\quad\{"diagnosis": "<exact dx name from list>", "update": <-2|-1|0|1|2>,
"rationale": "<one sentence>"\}\newline
\quad],\newline
\quad"ranked\_differential": ["<most likely>", "<second>", ..., "<least likely>"],\newline
\quad"reasoning\_summary": "<one sentence overall reasoning>"\newline
\}\newline
\newline
Update score semantics:\newline
+2 = this evidence strongly increases the likelihood of this diagnosis\newline
+1 = this evidence mildly increases the likelihood of this diagnosis\newline
\phantom{+}0 = this evidence does not meaningfully change the likelihood\newline
-1 = this evidence mildly decreases the likelihood of this diagnosis\newline
-2 = this evidence strongly decreases the likelihood of this diagnosis\newline
\newline
FINAL STEP ONLY --- also include:\newline
\quad"final\_diagnosis": "<single most likely diagnosis from the fixed list>"
\end{quote}

\paragraph{Per-Step Prompt — Baseline (user turn).}
\begin{quote}
\ttfamily
=== PATIENT CASE ===\newline
\{one\_sentence\}\newline
\newline
=== FIXED DIAGNOSIS LIST (re-rank these at every step) ===\newline
- \{dx\_1\}\newline
- \{dx\_2\}\newline
\quad\ldots\newline
\newline
=== NEW EVIDENCE (Step \{timestep\}) ===\newline
\{evidence\}\newline
\newline
=== DIAGNOSIS THIS EVIDENCE RELATES TO ===\newline
\{current\_dx\}\newline
\newline
=== YOUR PREVIOUS RANKING (Step \{timestep - 1\}) ===\newline
1. \{dx\_ranked\_1\}\newline
2. \{dx\_ranked\_2\}\newline
\quad\ldots\newline
\newline
Given the evidence above, update your belief and re-rank the full list.
\end{quote}

\noindent\textit{Note:} The previous ranking section is omitted at timestep 1,
replaced with: \texttt{"This is the first piece of evidence. Provide your initial
belief score and rank the full list."} At the final timestep the suffix
\texttt{"This is the FINAL step. Also provide `final\_diagnosis' --- your single best
diagnosis from the fixed list."} is appended.

\subsection*{A.5\quad SCT Clinical Reasoning — Single Oracle}

The system prompt is identical to Section~A.4.
The per-step user prompt is identical to the Baseline (Section~A.4)
with the following section inserted immediately before the previous-ranking block:

\begin{quote}
\ttfamily
=== PHYSICIAN RATIONALE FOR THIS STEP ===\newline
\{current\_rationale\}
\end{quote}

\noindent where \texttt{current\_rationale} is the one-sentence physician rationale
associated with the current \texttt{(encounterkey, differential)} pair from the
ground-truth annotation table.

\subsection*{A.6\quad SCT Clinical Reasoning — Full Oracle}

The system prompt is identical to Section~A.4.
The per-step user prompt is identical to the Baseline (Section~A.4)
with \emph{two} additional sections inserted immediately before the
previous-ranking block:

\begin{quote}
\ttfamily
=== PHYSICIAN RATIONALE FOR PRIOR STEPS ===\newline
Step \{t\_1\} (\{dx\_1\}): \{rationale\_1\}\newline
Step \{t\_2\} (\{dx\_2\}): \{rationale\_2\}\newline
\quad\ldots\newline
\newline
=== PHYSICIAN RATIONALE FOR THIS STEP ===\newline
\{current\_rationale\}
\end{quote}

\noindent At timestep 1 the prior-steps section is absent (no prior rationales exist).
At timestep $T$, rationales for all steps $1, \ldots, T$ are included in order.
Each prior rationale entry is drawn from the ground-truth annotation table matched
on \texttt{encounterkey} and \texttt{differential}.

\subsection*{A.7\quad Clinical Knowledge}

This experiment presents the model with the full patient summary and
all differential/evidence pairs simultaneously, then asks for a single
final diagnosis. Unlike the sequential SCT conditions
(Sections~A.4--A.6), no step-by-step belief updating is performed.
 
\paragraph{System Prompt.}
\begin{verbatim}
You are an experienced emergency medicine physician.
You will be given a patient presentation and a list of diagnoses to
consider, along with supporting clinical evidence for each.
Your task is to identify the single most likely diagnosis exactly how
it is written from the provided list.
Respond with ONLY the diagnosis name. Do not include any explanation,
punctuation, or additional text.
\end{verbatim}
 
\paragraph{Per-Encounter Prompt (user turn).}
\begin{verbatim}
=== PATIENT CASE ===
{one_sentence}
 
=== DIAGNOSES TO CONSIDER ===
- {dx_1}
- {dx_2}
...
 
=== CLINICAL EVIDENCE ===
Evidence 1 (related to {dx_1}): {evidence_1}
Evidence 2 (related to {dx_2}): {evidence_2}
...
 
=== TASK ===
Based on the patient case and all clinical evidence above, what is the
single most likely diagnosis?
You MUST choose exactly one diagnosis from the list above. Respond with
ONLY the diagnosis name, nothing else.
\end{verbatim}
 
\noindent The diagnosis list and evidence items are drawn from all
populated \texttt{differential\_\{t\}} / \texttt{evidence\_\{t\}} pairs
($t = 1, \ldots, 5$) for each encounter. 
 


\begin{table*}[ht]
\centering
\caption{Model configurations used across all experiments. Max output
tokens are reported as a range across tasks
($\text{min} \leq \cdot \leq \text{max}$), where the lower bound applies
to classification tasks (acuity, disposition, final diagnosis) and the
upper bound applies to the SCT sequential reasoning task. Reasoning
models require a larger budget to accommodate internal chain-of-thought
tokens before the final answer is emitted; using the lower bound for SCT
risks truncation. N/A\,=\,parameter not supported by that model or API.}
\label{tab:model_configs}
\footnotesize
\setlength{\tabcolsep}{5pt}
\begin{tabular}{lllcc>{\raggedright\arraybackslash}p{3.8cm}}
\toprule
\textbf{Model} & \textbf{Version / Snapshot} & \textbf{API} & \textbf{Temp.}
  & \textbf{Max Output Tokens} & \\
\midrule
\multicolumn{6}{l}{} \\
\midrule
DeepSeek-R1
  & \texttt{deepseek/deepseek-r1}
  & OpenRouter & 0.1
  & $2{,}000 \leq \cdot \leq 10{,}000$
  & \\[2pt]
o4-mini
  & \texttt{o4-mini-2025-04-16}
  & Azure OpenAI & N/A
  & $2{,}000 \leq \cdot \leq 10{,}000$
  & \\[2pt]
Gemini 2.5 Flash
  & \texttt{gemini-2.5-flash}
  & OpenRouter & 0.1
  & $100 \leq \cdot \leq 1{,}500$
  & \\[2pt]
GPT-5.2 Thinking
  & \texttt{gpt-5.2-2025-12-11}
  & Azure OpenAI & 0.1
  & $2{,}000 \leq \cdot \leq 10{,}000$
  & \\[2pt]
Claude Sonnet 4.5 Thinking
  & \texttt{claude-sonnet-4-5-20250929}
  & OpenRouter & N/A
  & $2{,}000 \leq \cdot \leq 10{,}000$
  & \\[2pt]
Phi-4
  & \texttt{microsoft/phi-4}
  & OpenRouter & 0.1
  & $100 \leq \cdot \leq 1{,}500$
  & \\[2pt]
\midrule
\multicolumn{6}{l}{} \\
\midrule
GPT-5.2
  & \texttt{gpt-5.2-2025-12-11}
  & Azure OpenAI & 0.1
  & $100 \leq \cdot \leq 1{,}500$
  & \\[2pt]
Claude Sonnet 4.5
  & \texttt{claude-sonnet-4-5-20250929}
  & OpenRouter & N/A
  & $100 \leq \cdot \leq 1{,}500$
  & \\[2pt]
\bottomrule
\end{tabular}
\end{table*}
\begin{table}[h]
\centering
\caption{Ground truth acuity and disposition distribution in ER-REASON. The cross-stage analysis (n=3,285) retains only encounters where acuity and disposition are concordant: high-acuity patients who were admitted and low-acuity patients who were discharged. Excluded encounters (n=685) include high-acuity patients discharged (n=434), low-acuity patients admitted (n=35), ESI 3 patients with transfer or other dispositions (n=112), transfer/other dispositions for high/low acuity (n=104), and unspecified acuity (n=14).}
\label{tab:base_rates}
\begin{tabular}{lrrrrr}
\toprule
\textbf{Acuity Group} & \textbf{Admit} & \textbf{Discharge} & \textbf{Transfer} & \textbf{Other} & \textbf{Total} \\
\midrule
High (ESI 1--2)  & 1,110 & 434 & 74 & 21 & 1,639 \\
Mid (ESI 3)      & 1,148 & 920 & 69 & 43 & 2,180 \\
Low (ESI 4--5)   &    35 & 107 &  6 &  3 &   151 \\
\midrule
\textbf{Total}   & \textbf{2,293} & \textbf{1,461} & \textbf{149} & \textbf{67} & \textbf{3,970} \\
\midrule
\multicolumn{6}{l}{\textit{Retained in cross-stage analysis (concordant only)}} \\
\midrule
High bucket (ESI 1--2 + ESI 3 admitted)    & 2,258 & -- & -- & -- & 2,258 \\
Low bucket (ESI 4--5 + ESI 3 discharged)   & -- & 1,027 & -- & -- & 1,027 \\
\midrule
\textbf{Analysis total} & \textbf{2,258} & \textbf{1,027} & -- & -- & \textbf{3,285} \\
\bottomrule
\end{tabular}%
\end{table}






\newpage
\section*{NeurIPS Paper Checklist}

\begin{enumerate}

\item {\bf Claims}
    \item[] Question: Do the main claims made in the abstract and introduction accurately reflect the paper's contributions and scope?
    \item[] Answer: \answerYes{} 
    \item[] Justification: The main claims in the paper accurately reflect the paper’s contributions and scope. The paper introduces ER-Reason, a benchmark designed to evaluate large language models (LLMs) on realistic, high-stakes clinical reasoning tasks in the emergency room (ER).

\item {\bf Limitations}
    \item[] Question: Does the paper discuss the limitations of the work performed by the authors?
    \item[] Answer:  \answerYes{} 
    \item[] Justification: The limitations are explicitly acknowledged including the scope of clinical coverage, specifically how our tasks do not account for hospital-level contextual factors, such as bed availability, staffing constraints, and institutional protocols, which can significantly influence decisions such as triaging and patient disposition. We also acknowledge how we did not control for inference-time thinking budget for reasoning models.

\item {\bf Theory assumptions and proofs}
    \item[] Question: For each theoretical result, does the paper provide the full set of assumptions and a complete (and correct) proof?
    \item[] Answer: \answerNA{}
    \item[] Justification: The paper does not have theoretical claims or results that would require a set of assumptions or proofs.

    \item {\bf Experimental result reproducibility}
    \item[] Question: Does the paper fully disclose all the information needed to reproduce the main experimental results of the paper to the extent that it affects the main claims and/or conclusions of the paper (regardless of whether the code and data are provided or not)?
    \item[] Answer: \answerYes{}
    \item[] Justification: Our medical benchmark is hosted on PhysioNet, and requires the highest credentialed access in accordance with our University’s Compliance team. We also upload our code to a public GitHub repository. 

\item {\bf Open access to data and code}
    \item[] Question: Does the paper provide open access to the data and code, with sufficient instructions to faithfully reproduce the main experimental results, as described in supplemental material?
    \item[] Answer: \answerYes{}
    \item[] Justification: We release open access to all code and instructions necessary to reproduce the core experimental results. As mentioned, our data is released on PhysioNet since it is de-identified patient data and requires the highest credentialed access.

\item {\bf Experimental setting/details}
    \item[] Question: Does the paper specify all the training and test details (e.g., data splits, hyperparameters, how they were chosen, type of optimizer) necessary to understand the results?
    \item[] Answer:\answerYes{}
    \item[] Justification: Our study exclusively uses prompting-based inference on large language models (LLMs). We provide detailed descriptions of the prompt templates used for each task in the Appendix. 

\item {\bf Experiment statistical significance}
    \item[] Question: Does the paper report error bars suitably and correctly defined or other appropriate information about the statistical significance of the experiments?
    \item[] Answer: \answerYes{}
    \item[] Justification: We report confidence intervals for all of our results. 

\item {\bf Experiments compute resources}
    \item[] Question: For each experiment, does the paper provide sufficient information on the computer resources (type of compute workers, memory, time of execution) needed to reproduce the experiments?
    \item[] Answer: \answerYes{}
    \item[] Justification: All experiments were conducted using pre-trained LLMs accessed via the OpenRouter API. No dedicated compute resources (e.g., GPUs, distributed workers, or specialized memory requirements) are needed to reproduce the experiments locally
    
\item {\bf Code of ethics}
    \item[] Question: Does the research conducted in the paper conform, in every respect, with the NeurIPS Code of Ethics \url{https://neurips.cc/public/EthicsGuidelines}?
    \item[] Answer: \answerYes{}
    \item[] Justification: The patient data used are fully de-identified, and comply with university and institutional privacy policies. Ethical approvals and data use agreements were obtained, ensuring that patient privacy and confidentiality are strictly maintained throughout the research. 

\item {\bf Broader impacts}
    \item[] Question: Does the paper discuss both potential positive societal impacts and negative societal impacts of the work performed?
    \item[] Answer: \answerYes{} 
    \item[] Justification: The paper discusses the benefits and limitations of this benchmark. We advocate for a better definition of clinical reasoning, and model evaluation that is grounded in real-world EHR data. At the same time, we acknowledge the limitations of our dataset such as variability clinical documentation, challenges in exact diagnostic coding, and the benchmark’s focus on a specific clinical setting. 
    
\item {\bf Safeguards}
    \item[] Question: Does the paper describe safeguards that have been put in place for responsible release of data or models that have a high risk for misuse (e.g., pre-trained language models, image generators, or scraped datasets)?
    \item[] Answer: \answerYes{}
    \item[] Justification: The paper details multiple safeguards implemented to ensure the responsible release and use of data. Any access to de-identified clinical data is restricted to users with the highest-level credentials and institutional approvals on PhysioNet, and must have undergone CITI training. Data handling strictly follows university compliance and HIPAA regulations. These measures align with best practices for ethical AI deployment in medicine.

\item {\bf Licenses for existing assets}
    \item[] Question: Are the creators or original owners of assets (e.g., code, data, models), used in the paper, properly credited and are the license and terms of use explicitly mentioned and properly respected?
    \item[] Answer: \answerYes{}
    \item[] Justification: All data and code used in this paper are either developed in-house or or obtained through institutional resources followed appropriate protocol. 

\item {\bf New assets}
    \item[] Question: Are new assets introduced in the paper well documented and is the documentation provided alongside the assets?
    \item[] Answer:\answerYes{}
    \item[] Justification: We introduce a new medical benchmark dataset and evaluation framework for clinical reasoning. The benchmark is fully documented, and all associated data and code are released with detailed instructions to facilitate reproducibility and use by future researchers. 

\item {\bf Crowdsourcing and research with human subjects}
    \item[] Question: For crowdsourcing experiments and research with human subjects, does the paper include the full text of instructions given to participants and screenshots, if applicable, as well as details about compensation (if any)? 
    \item[] Answer: \answerYes{}
    \item[] Justification: The study involves physician annotators who performed expert labeling of medical data. We describe the annotation procedure in the supplementary material.

\item {\bf Institutional review board (IRB) approvals or equivalent for research with human subjects}
    \item[] Question: Does the paper describe potential risks incurred by study participants, whether such risks were disclosed to the subjects, and whether Institutional Review Board (IRB) approvals (or an equivalent approval/review based on the requirements of your country or institution) were obtained?
    \item[] Answer: \answerYes{}
    \item[] Justification: We’ve taken appropriate measures to disclose risks to participants via a university Institutional Review Board (IRB). 

\item {\bf Declaration of LLM usage}
    \item[] Question: Does the paper describe the usage of LLMs if it is an important, original, or non-standard component of the core methods in this research? Note that if the LLM is used only for writing, editing, or formatting purposes and does \emph{not} impact the core methodology, scientific rigor, or originality of the research, declaration is not required.
    \item[] Answer: \answerYes{}
    \item[] Justification: The paper clearly describes the use of large language models (LLMs) as a central part of the methodology. Usage details and models are well-documented throughout our paper. 

\end{enumerate}

\end{document}